\documentclass[conference]{IEEEtran}
\IEEEoverridecommandlockouts
\usepackage{cite}
\usepackage{amsmath,amssymb,amsfonts}
\usepackage{algorithmic}
\usepackage{graphicx}
\usepackage{textcomp}
\usepackage{xcolor}

\usepackage{witharrows}

\usepackage[flushleft]{threeparttable}
\usepackage{tablefootnote}
\usepackage{array}
\newcolumntype{P}[1]{>{\centering\arraybackslash}p{#1}}
\usepackage{multicol}
\usepackage{multirow}
\usepackage{tabulary}
\usepackage{colortbl}
\definecolor{mygray}{gray}{0.90}
\usepackage{makecell}
\usepackage{booktabs}
\usepackage{subcaption}

\usepackage{stmaryrd}
\usepackage{mathabx}
\usepackage{float}

\usepackage{url,hyperref,microtype}
\usepackage[switch,columnwise]{lineno}
\def\BibTeX{{\rm B\kern-.05em{\sc i\kern-.025em b}\kern-.08em
    T\kern-.1667em\lower.7ex\hbox{E}\kern-.125emX}}

\usepackage{fancyhdr}
\begin{document}

\title{Twins-PainViT: Towards a Modality-Agnostic Vision Transformer Framework for Multimodal Automatic Pain Assessment using Facial Videos and fNIRS\\
\thanks{$\dagger$: Corresponding Author, $\ddagger$: Affiliated Researcher at the Computational
Biomedicine Laboratory of the Foundation for Research and Technology
(FORTH), Heraklion, Greece }
}

\author{\IEEEauthorblockN{Stefanos Gkikas$^\dagger$}
\IEEEauthorblockA{\textit{Department of Electrical \& Computer Engineering} \\
\textit{Hellenic Mediterranean University}\\
Heraklion, Greece\\
gkikas@ics.forth.gr}
\and
\IEEEauthorblockN{Manolis Tsiknakis$^\ddagger$}
\IEEEauthorblockA{\textit{Department of Electrical \& Computer Engineering} \\
\textit{Hellenic Mediterranean University}\\
Heraklion, Greece \\
tsiknaki@ics.forth.gr}

}

\maketitle

\begin{abstract}
Automatic pain assessment plays a critical role for advancing healthcare and optimizing pain management strategies. This study has been submitted to the \textit{First Multimodal Sensing Grand Challenge for Next-Gen Pain Assessment (AI4PAIN)}.
The proposed multimodal framework utilizes facial videos and fNIRS and presents a modality-agnostic approach, alleviating the need for domain-specific models.
Employing a dual ViT configuration and adopting waveform representations for the fNIRS, as well as for the extracted embeddings from the two modalities, demonstrate the efficacy of the proposed method, achieving an accuracy of $46.76\%$ in the multilevel pain assessment task.

\end{abstract}


\section{Introduction}
The International Association for the Study of Pain (IASP) defines \textit{pain} as \textit{\textquotedblleft an unpleasant sensory and emotional experience associated with actual or potential tissue damage, or described in terms of such damage\textquotedblright} \cite{iasp_2020}, marking a recent update to the definition.
Pain significantly affects individuals and societal structures, with people of all ages experiencing it due to accidents, diseases, or medical treatments—making it the primary reason for medical consultations. Acute and chronic pain pose clinical, economic, and social difficulties. Beyond its direct effects on a person's daily life, pain is associated with various negative consequences, such as increased opioid use, substance abuse, addiction, declining social interactions, and mental health problems \cite{dinakar_stillman_2016}.
Effective pain assessment is essential for early diagnosis, disease progression monitoring, and evaluation of treatment efficacy, especially in managing chronic pain \cite{gkikas_tsiknakis_slr_2023}.
Additionally, adjusting pain intensity is crucial in therapy approaches like myofascial therapy, where a practitioner, such as a physiotherapist, externally induces the pain, and understanding the patient's pain level is vital \cite{badura_2021}.
Pain evaluation is crucial yet challenging for healthcare professionals \cite{aqajari_cao_2021}, especially when dealing with patients who cannot communicate verbally. This challenge is further amplified in elderly patients who may be less expressive or hesitant to discuss their pain \cite{yong_gibson_2001}. Moreover, comprehensive research \cite{bartley_fillingim_2013,gkikas_chatzaki_2022, gkikas_chatzaki_2023} highlights significant differences in pain expression across different genders and age groups, adding complexity to the pain assessment process.
Pain assessment encompasses a variety of approaches, from self-reporting using detailed rating scales and questionnaires, considered the gold standard, to observing behavioral indicators like facial expressions, vocalizations, and bodily movements \cite{rojas_brown_2023}. 
It also includes analyzing physiological responses such as electrocardiography and skin conductance, which offer essential insights into the physical manifestations of pain \cite{gkikas_tsiknakis_slr_2023}. Moreover,
functional near-infrared spectroscopy (fNIRS) is a promising method for measuring pain-related physiological responses. This non-invasive neuroimaging technique evaluates brain activity by tracking cerebral hemodynamics and oxygenation changes. Specifically, fNIRS simultaneously records changes in the cortical concentrations of oxygenated hemoglobin (HbO) and deoxygenated hemoglobin (HbR), offering critical insights into brain function \cite{rojas_huang_2016}.
Furthermore, fNIRS studies have demonstrated that noxious stimuli initiate changes in oxygenation levels across various cortical regions in healthy and diseased subjects \cite{rojas_liao_2019}.

This study introduces a modality-agnostic multimodal framework that utilizes videos and fNIRS. The proposed pipeline is based on a dual Vision Transformer (ViT) configuration, eliminating the need for domain-specific architectures or extensive feature engineering for each modality by interpreting the inputs as unified images through 2D waveform representation.

\section{Related Work}
\label{related_work}
Recent developments have introduced various innovative methods for assessing pain levels from video data.
The authors in \cite{bargshady_zhou_2020} developed a temporal convolutional network (TCN) and utilized the HSV color model, arguing that it offers more advantages for tasks related to human visual perception, such as skin pixel detection and multi-face detection.
The authors in \cite{bargshady_soar_2019} combined the VGG-Face CNN with a 3-layer LSTM to extract spatio-temporal features from grayscale images, applying zero-phase component analysis for enhancement. Conversely, in \cite{bargshady_zhou_2020_b}, principal component analysis was employed to reduce dimensionality.
Finally, in \cite{gkikas_tsiknakis_embc}, the authors introduced a hybrid approach that combines a vision transformer for spatial feature extraction with a standard transformer for temporal analysis, achieving high accuracy.
Several studies in pain research have employed fNIRS in conjunction with machine learning techniques to extract relevant features and evaluate pain conditions effectively.
In \cite{rojas_huang_2017_b}, combining a bag-of-words (BoW) approach with a K-NN classifier to analyze time-frequency features yields better results than analyzing time or frequency features in isolation.
Conversely, the study \cite{rojas_juang_2019} demonstrated that the best results were achieved by combining time and frequency domain features with a Gaussian SVM, while Rojas \textit{et al}. \cite{rojas_romero_2021} utilized the raw fNIRS with a two-layer BiLSTM achieving $90.60\%$ accuracy in a 
multi-class classification task. 
Finally, the authors in \cite{rojas_joseph_2024} developed a hybrid architecture of CNN and LSTM model to capture spatio-temporal features from the fNIRS, achieving high performances. 
Regarding the multimodal approaches, Gkikas \textit{et al}. \cite{gkikas_tachos_2024} introduced an efficient transformer-based multimodal framework that leverages facial videos and heart rate signals, demonstrating that integrating behavioral and physiological modalities enhances pain estimation performance.
In \cite{rojas_hirachan_2023}, statistical features were extracted from electrodermal activity, respiration rate, and photoplethysmography, and a joint mutual information process was implemented to assess the intensity and locate the origin of pain. 

\section{Methodology}
This section describes the pipeline of the proposed multimodal automatic pain assessment framework, the architecture of the models, the pre-processing methods, the pre-training strategy, and the  augmentation techniques.

\subsection{Framework Architecture}
The proposed framework, \textit{Twins-PainViT}, consists of two models: \textit{PainViT--1} and \textit{PainViT--2}. Both models are identical in architecture and parameters and follow the same pre-training procedure, which will be detailed in Section \ref{pretraining}.
\textit{PainViT--1} is provided with the corresponding video frames and the visualized fNIRS channels and functions as an embedding extractor. \textit{PainViT--2} receives the visual representation of the embeddings and completes the final pain assessment.

\subsubsection{PainViT}
Vision Transformers (ViTs) \cite{vit} have emerged as a new paradigm in computer vision tasks due to their performance. However, despite their impressive efficacy, transformer-based models face challenges in scaling with larger input sizes, leading to substantial computational costs. This inefficiency primarily derives from the element-wise operations in the multi-head self-attention mechanism.
Numerous efforts have been made to enhance the efficiency and reduce the complexity of transformer-based architectures by modifying the self-attention module or the model's overall structure \cite{mobile_former}\cite{pale_transformer}. Our approach is founded on the principles of \cite{swin_transformer} introducing the hierarchical architectures into the vision transformers and \cite{efficientvit} proposing mechanisms that increase efficiency and speed.

\subsubsection{PainViT--block}
Each block features two components: the \textit{Token-Mixer} and the \textit{Cascaded-Attention}. It is structured with an \textit{Cascaded-Attention} module at the core and a \textit{Token-Mixer} module positioned preceding and following it.
For every input image $I$, overlapping patch embedding is applied, producing $16\times 16$ patches, each projected into a token with a dimension of $d$. 
\paragraph{Token-Mixer}
To enhance the incorporation of local structural information, the token $T$ is processed through a depthwise convolution layer:
\begin{equation}
Y_{c} = K_c * T_{c} + b_c,
\end{equation}
where $Y_{c}$ is the output of the depthwise convolution for channel $c$ of the token $T_c$.
$K_c$ is the convolutional kernel specifically for channel $c$, 
$T_{c}$ is the $c$-th channel of the token $T$, and
$b_c$ is the bias term added to the convolution output of channel $c$.
The symbol $*$ denotes the convolution operation.
Following the depthwise convolution, batch normalization is applied to the output:
\begin{equation}
Z_{c} = \gamma_c \left( \frac{Y_{c} - \mu_B}{\sqrt{\sigma_B^2 + \epsilon}} \right) + \beta_c,
\end{equation}
where $Z_{c}$ is the batch-normalized output for channel $c$ of the token $T$.
$\gamma_c$ and $\beta_c$ are learnable parameters specific to channel $c$ that scale and shift the normalized data.
$\mu_B$ is the batch mean of $Y_{c}$,
$\sigma_B^2$ is the batch variance of $Y_{c}$,
and $\epsilon$ is a small constant added for numerical stability to avoid division by zero.
Next, a feed-forward network (FFN) facilitates more efficient communication between different feature channels:
\begin{equation}
\Phi^F(Z_{c}) = W_2 \cdot \text{ReLU}(W_1 \cdot Z_{c} + b_1) + b_2,
\end{equation}
where $\Phi^F(Z_{c})$ is the output of the feed-forward network for the input $Z_{c}$.
$W_1$ and $W_2$ are the weight matrices of the first and second linear layers;
$b_1$ and $b_2$ are the bias terms for the first and second linear layers, respectively,
and $\text{ReLU}$ is the activation function.

\paragraph{Cascaded-Attention}
Regarding the attention mechanism, there is a single self-attention layer.
For every input embedding:
\begin{equation}
X_{i+1}=\Phi^A(X_i),
\end{equation}
where $X_i$ is the full input embedding for the $i$-th \textit{PainViT-block}.
More specifically, the \textit{Cascaded-Attention} module employs a cascaded mechanism that partitions the full input embedding into smaller segments, each directed to a distinct attention head. This approach allows the computation to be distributed across the heads, enhancing efficiency by avoiding long input embeddings.
The attention is described as: 
\begin{equation}
\widetilde{X}_{ij} = Attn({X}_{ij} W^Q_{ij}, {X}_{ij} W^K_{ij}, {X}_{ij} W^V_{ij}),
\end{equation}
\begin{equation}
\widetilde{X}_{i+1} = Concat[\widetilde{X}_{ij}]_{j=1:h}W^P_i,
\end{equation}
where each $j$-th head calculates the self-attention for $X_{i,j}$, which represents the $j$-th segment of the full input embedding $X_i$, structured as $[X_{i1}, X_{i2}, \dots, X_{ih}]$ where $1 \leq j \leq h$ and $h$ denotes the total number of heads. The projection layers $W^Q_{ij}$, $W^K_{ij}$, and $W^V_{ij}$ map each segment input embedding into distinct subspaces. Finally, $W^P_i$ is a linear layer that reassembles the concatenated output embeddings from all heads back to a dimensionality that aligns with the original input.
Furthermore, the cascaded architecture enhances the learning of richer embedding representations for $Q$, $K$, and $V$ layers. This is achieved by adding the output from each head to the input of the subsequent head, enabling the accumulation of information throughout the process. Specifically: 
\begin{equation}
X^{'}_{ij} = X_{ij} + \widetilde{X}_{i(j-1)}.
\end{equation}
Here, $X'_{ij}$ represents the addition of the $j$-th input segment $X_{ij}$ and the output $\tilde{X}_{i(j-1)}$ from the 
$(j-1)$-th head. The summation replaces $X_{ij}$ as the new input embedding for the
$j$-th head in the self-attention computation.
Finally, it is noted that depthwise convolution is applied to each $Q$ in every attention head. This enables the subsequent self-attention process to capture global representations and local information.

\begin{figure*}
\begin{center}
\includegraphics[scale=0.57]{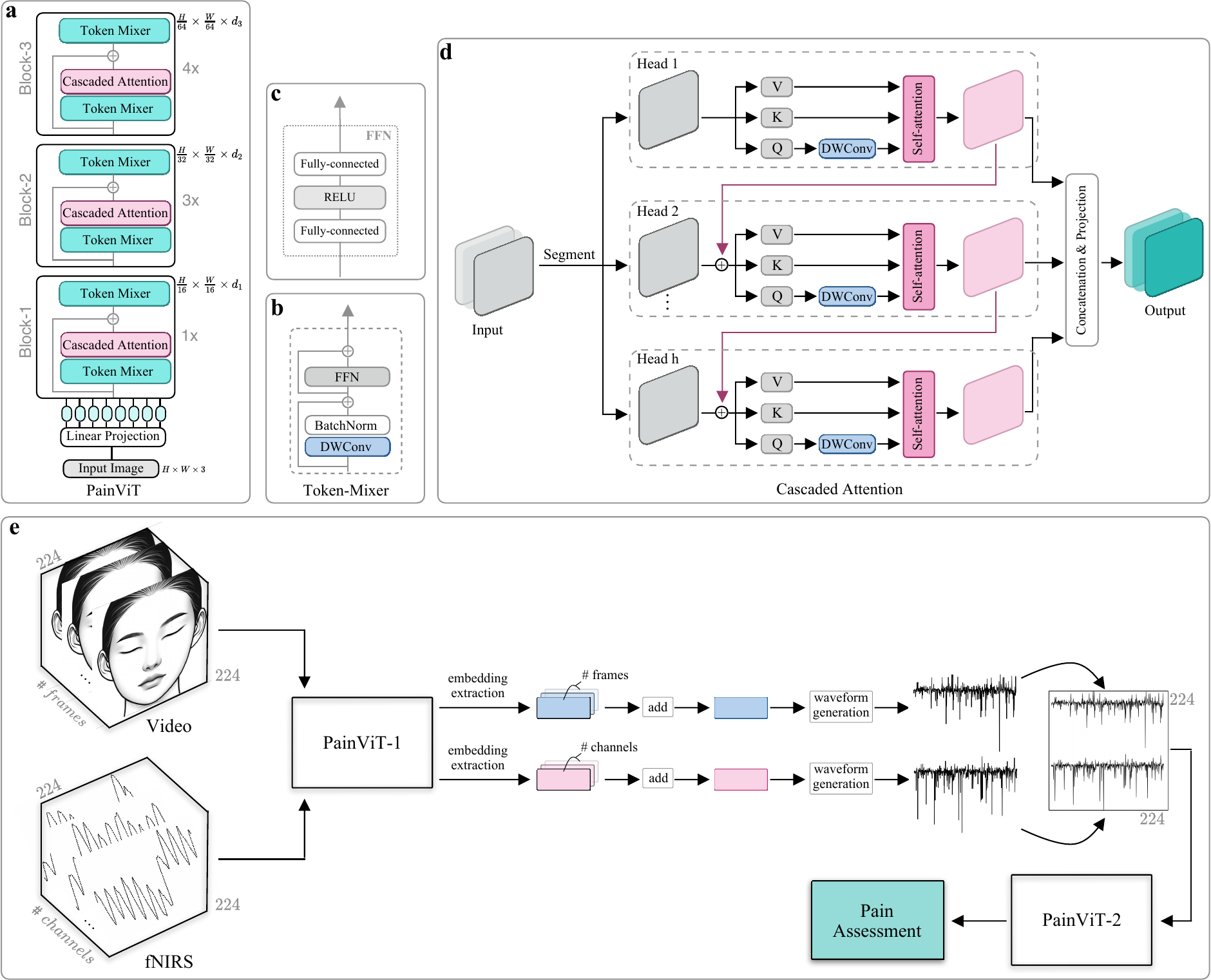} 
\end{center}
\caption{\textit{PainViT}: \textbf{(a)} Hierarchical organization of the \textit{PainViT} blocks, each with different depths, illustrating the reduction in token resolution at each stage; \textbf{(b)} Detail of the Token-Mixer module, showcasing its components including a depthwise convolution (DWConv) and batch normalization; \textbf{(c)} The Feed-Forward Network (FFN) structure within the \textit{Token-Mixer}; \textbf{(d)} The \textit{Cascaded Attention} mechanism across multiple heads, depicting the process of adding outputs from previous heads to enhance the self-attention computation, and the final output projection; \textbf{(e)} Overview of the proposed multimodal pipeline, utilizing videos and fNIRS. The extracted embeddings from \textit{PainViT--1} are visualized as waveform diagrams, which are then combined into a single diagram depicting both modalities before being entered into \textit{PainViT--2} for the final pain assessment.}
\label{full}
\end{figure*}

The framework comprises three \textit{PainViT--blocks}, each with $1$, $3$, and $4$ depths, respectively. This hierarchical structure features a progressive reduction in the number of tokens by subsampling the resolution by a factor of $2\times$ at each stage. Correspondingly, the architecture facilitates the extraction of embeddings with dimensions $d$ across the blocks, specifically $192$, $288$, and $500$. Additionally, the multihead self-attention mechanism within each block employs $3$, $3$, and $4$ heads, respectively.
\hyperref[full]{Fig. 1(a-d}) illustrates the \textit{PainViT} architecture and its fundamental building blocks, while Table \ref{table:module_parameters} presents the number of parameters and the computational cost in terms of floating-point operations (FLOPS).

\renewcommand{\arraystretch}{1.2}
\begin{table}
\caption{Number of parameters and FLOPS for the components of the proposed Twins-PainViT.}
\label{table:module_parameters}
\begin{center}
\begin{threeparttable}
\begin{tabular}{ P{3.3cm}  P{2.0cm}  P{2.0cm}}
\toprule
Module & Params (M) &FLOPS  (G) \\
\midrule
\midrule
PainViT--1   &16.46 &0.59\\
PainViT--2   &16.46 &0.59 \\
\hline
Total &32.92 &1.18\\
\bottomrule
\end{tabular}
\begin{tablenotes}
\scriptsize
\item 
\end{tablenotes}
\end{threeparttable}
\end{center}
\end{table}


\subsection{Embedding extraction \& Fusion}
For each frame of a video, $V = [v_1, v_2, \ldots, v_n]$, \textit{PainViT--1} extracts a corresponding embedding. These embeddings are aggregated to form a unified feature representation of the video. Similarly, for each channel of an fNIRS signal, $C = [c_1, c_2, \ldots, c_m]$, \textit{PainViT--1} extracts embeddings, which are subsequently aggregated to create a representation of the fNIRS signal. This process can be described as:  
\begin{equation}
E_V \leftarrow \sum_{i=1}^n \text{\textit{PainViT--1}}(v_i),
\end{equation}

\begin{equation}
E_C \leftarrow \sum_{i=1}^m \text{\textit{PainViT--1}}(c_i),
\end{equation}
where $E_V$ and $E_C$ are the corresponding embedding representations for the video and fNIRs. 
Following the extraction of embeddings, $E_V$ and $E_C$ are visualized as waveform diagrams. 
The waveform from each modality—video and fNIRS—is merged into a single image with a resolution of $224\times 224$. This unified visual representation is fed into \textit{PainViT--2} for the final pain assessment.
(\hyperref[full]{Fig. 1e}) presents a high-level overview of the multimodal proposed pipeline.


\subsection{Pre-processing}
\label{preprocessing}
\begin{figure}
\begin{center}
\includegraphics[scale=0.275]{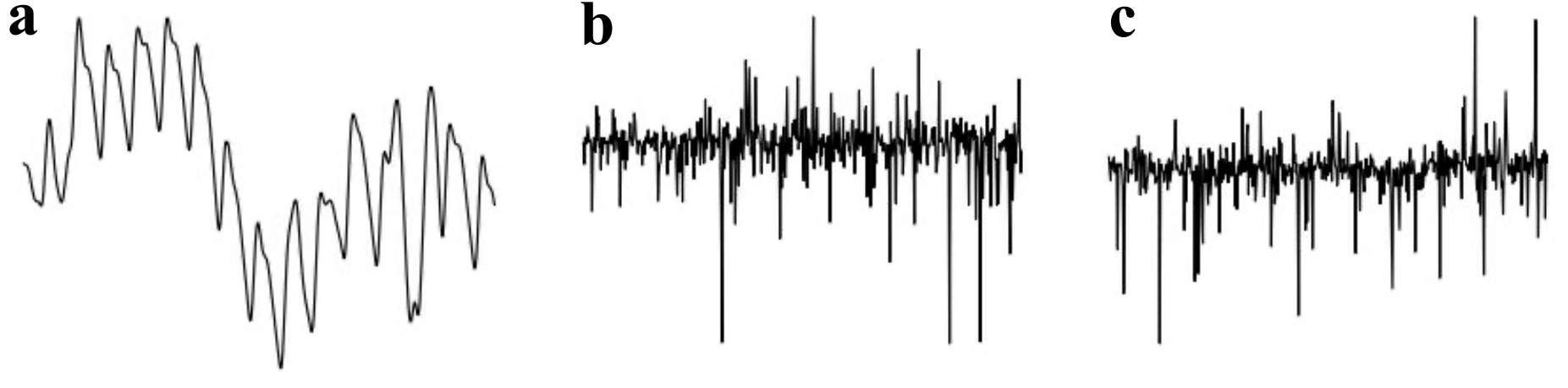}
\end{center}
\caption{Waveform diagrams representing different data modalities: \textbf{(a)} original fNIRS signal waveform, \textbf{(b)} video embedding extracted from \textit{PainViT--1}, and \textbf{(c)} fNIRS embedding extracted from \textit{PainViT--1}. }
\label{waveforms}
\end{figure}

The pre-processing involves face detection for the corresponding frames of the videos and the generation of waveform diagrams from the original fNIRS. 
The MTCNN face detector \cite{zhang_2016} was utilized, employing a series of cascaded convolutional neural networks to predict both faces and facial landmarks.
The resolution of the detected faces was set at $224\times 224$ pixels.  
All fNIRS channels are used to generate waveform diagrams. A waveform diagram visually represents the shape and form of a signal wave as it progresses over time, illustrating the signal's amplitude, frequency, and phase. This method provides the simplest and most direct way to visualize a signal, as it does not necessitate any transformations or additional computations such as those involved in creating spectrograms, scalograms, or recurrence plots.
Similarly, the embeddings extracted from \textit{PainViT--1} are visualized using the same method. Although these embeddings are not signals, the 1D vectors can still be plotted in a 2D space for analysis or utilization from the deep-learning vision models. 
All waveform diagrams generated from the fNIRS data and embeddings are formatted as images with a $224\times 224$ pixels resolution. Fig. \ref{waveforms} depicts waveform representations of channel-specific fNIRS signal, an embedding extracted from a video, and an embedding derived from a channel-specific fNIRS sample.

\subsection{Pre-training}
\label{pretraining}
Before the automatic pain assessment training process, the \textit{Twins-PainViT} models were pre-trained using a multi-task learning strategy. Four datasets, which include images for emotion assessment tasks, were employed. The \textit{AffectNet} \cite{mollahosseini_hasani_2019} and \textit{RAF-DB basic} \cite{li_deng_2017} datasets provided facial images for recognizing basic emotions, while the \textit{Compound FEE-DB} \cite{du_tao_2014} and \textit{RAF-DB compound} \cite{li_deng_2017} datasets were used for identifying complex emotions.
Additionally, five datasets containing biosignals were also utilized. 
\textit{EEG-BST-SZ} \cite{ford_2013} comprises electroencephalograms used for schizophrenia recognition, and \textit{Silent-EMG} \cite{gaddy_klein_2020} includes electromyograms aimed at identifying the location of origin of the EMGs (such as throat and mid-jaw). Furthermore, electrocardiogram, electromyogram, and galvanic skin response samples from the \textit{BioVid} \cite{biovid_2013} dataset were employed for the pain assessment task.
All the biosignals were utilized in the form of waveform representations, as described in \ref{preprocessing}. 
The multi-task learning process is described as: 
\begin{equation}
L_{total} = \sum_{i=1}^9 \left[ e^{w_i} L_{S_i} + w_i \right],
\end{equation}
where $L_{S_i}$ is the loss associated with each specific task corresponding to different datasets, and $w_i$ represents the learned weights that guide the learning process in minimizing the overall loss $L_{total}$, taking into account all the individual losses.
Table \ref{table:datasets} details the datasets used in the pre-training process.

\renewcommand{\arraystretch}{1.2}
\begin{table}
\caption{Datasets utilized for the pretraining process of the framework.}
\label{table:datasets}
\begin{center}
\begin{threeparttable}
\begin{tabular}{ p{2.8cm} p{1.1cm} p{1.0cm} p{1.8cm} }
\toprule
Dataset &\#  samples &\# classes &Modality\\
\midrule
\midrule
\textit{AffectNet} \cite{mollahosseini_hasani_2019} &0.40M &8 &Facial Images\\
\textit{RAF-DB basic} \cite{li_deng_2017}&15,000 &7 &Facial Images\\
\textit{RAF-DB compound} \cite{li_deng_2017}&4,000 &11 &Facial Images\\
\textit{Compound FEE-DB}  \cite{du_tao_2014}&6,000 &26 &Facial Images\\
\textit{EEG-BST-SZ} \cite{ford_2013}&1.5M &2 &EEG\\
\textit{Silent-EMG} \cite{gaddy_klein_2020}&190,816 &8 &EMG\\
\textit{BioVid} \cite{biovid_2013}&8,700 &5 &ECG\\
\textit{BioVid} \cite{biovid_2013}&8,700 &5 &EMG\\
\textit{BioVid} \cite{biovid_2013}&8,700 &5 &GSR\\
 
\bottomrule 
\end{tabular}
\begin{tablenotes}
\scriptsize
\item EEG: electroencephalogram\space EMG: electromyogram\space ECG: electrocardiogram\space GSR: galvanic skin response 
\end{tablenotes}
\end{threeparttable}
\end{center}
\end{table}

\subsection{Augmentation Methods \& Regularization}
Several augmentation methods have been utilized for training the proposed framework. Regarding the pre-training process, \textit{RandAugment} \cite{randAugment} and \textit{TrivialAugment} \cite{trivialAugment} were adopted. Additionally, auxiliary noise from a uniform distribution was employed, along with \textit{MaskOut}, a custom-implemented technique masking out random square sections of input images.
For the automatic pain assessment task, \textit{AugMix} \cite{augmix} is employed in addition to \textit{RandAugment}, \textit{TrivialAugment}, and \textit{MaskOut} as augmentation methods.
Furthermore, \textit{Label Smoothing} \cite{label_smoothing} and \textit{DropOut} \cite{dropout} were employed as regularization techniques. 

\section{Experimental Evaluation \& Results}
This study utilizes the dataset provided by the challenge organizers \cite{ai4pain,rojas_hirachan_2023}, comprising facial videos and fNIRS data from $65$ participants. The dataset includes $41$ training, $12$ validation, and $12$ testing subjects recorded at the Human-Machine Interface Laboratory, University of Canberra, Australia.
Electrodes for transcutaneous electrical nerve stimulation, serving as pain stimuli, were placed on the inner forearm and the back of the right hand.
Pain threshold, defined as the lowest stimulus intensity at which stimulation becomes painful (low pain), and pain tolerance, defined as the highest intensity of pain a person can endure before it becomes intolerable (high pain), were measured.
For the fNIRS, $24$ channels each for HbO and HbR were utilized, alongside all $30$ frames per video available. The results presented in this study focus on the validation part of the dataset, structured in a multi-level classification setting (No Pain, Low Pain, and High Pain).
Table \ref{table:training_details} outlines the training framework details for the automatic pain assessment.
We note that numerous experiments were conducted across each modality and their fusion; however, only the most successful results are presented in the subsequent sections and corresponding tables.

\renewcommand{\arraystretch}{1.2}
\begin{table}
\caption{Training details for the automatic pain assessment.}
\label{table:training_details}
\begin{center}
\begin{threeparttable}
\begin{tabular}{ P{0.85cm} P{0.50cm}  P{0.70cm} P{0.75cm} P{0.6cm} P{0.75cm} P{0.9cm} P{0.5cm}}
\toprule
Optimizer & LR &LR decay &Weight decay &Epochs &Warmup epochs &Cooldown epochs &Batch size\\
\midrule
\midrule
\textit{AdamW}   &\textit{2e-5} &\textit{cosine}  &0.1 &100 &10 &10 &32\\
\bottomrule
\end{tabular}
\begin{tablenotes}
\scriptsize
\item LR: learning rate
\end{tablenotes}
\end{threeparttable}
\end{center}
\end{table}

\subsection{Facial Videos}
In the context of facial videos, two fusion embedding techniques were applied: the \textit{Addition} method aggregating the $30$ embeddings into a single fused vector with dimension $d=500$ and the \textit{Concatenation} method combining the embeddings to form a vector with $d=30\times 500=15,000$.
Utilizing the \textit{Addition} method, we observed an initial accuracy of $41.90\%$ for the multi-class classification task with augmentation and regularization levels ($0.1$ for \textit{AugMix}, \textit{Rand}, \textit{Trivial}, and $0.1|3$ for \textit{MaskOut}, and $0.5$ for \textit{DropOut}). Increasing the augmentation intensities to $0.5$ and \textit{MaskOut} to $0.7\vert3$ raised the accuracy to $44.91\%$. 
Applying \textit{MaskOut} to $0.7\vert3$ and raising \textit{DropOut} from $0.5$ to $0.6$ achieved $42.36\%$. Increasing \textit{DropOut} to $0.7$ and \textit{AugMix}, \textit{Rand}, and \textit{Trivial} to $0.9$ improved accuracy to $43.52\%$. 
Table \ref{table:videos_addition} presents the results.
Utilizing the \textit{Concatenation} method, and initial settings with a uniform augmentation probability of $0.3$ across \textit{AugMix}, \textit{Rand}, \textit{Trivial}, and $0.3\vert 3$ for \textit{MaskOut} and $0.1$ \textit{LS} and $0.5$ \textit{DropOut} yielded a $40.28\%$ accuracy. Increasing \textit{MaskOut} to $0.8\vert 5$ while maintaining other augmentations at $0.5$ improved accuracy to $41.44\%$. The highest accuracy of $43.75\%$ was achieved with $0.9$ across all augmentations except \textit{MaskOut}, which was adjusted to $0.6\vert 3$, and high regularization (\textit{LS} $0.4$, \textit{DropOut} $0.5$). The corresponding results are summarized in Table \ref{table:videos_concat}.

\renewcommand{\arraystretch}{1.2}
\begin{table}
\caption{Classification results utilizing the facial video modality \& \textit{Addition} method, reported on accuracy \%.}
\label{table:videos_addition}
\begin{center}
\begin{threeparttable}
\begin{tabular}{ P{0.8cm}  P{0.8cm} P{0.8cm} P{1.1cm}  P{0.5cm}  P{1.1cm}  P{0.8cm} }
\toprule
\multicolumn{4}{c}{Augmentation} 
&\multicolumn{2}{c}{Regularization} 
&\multicolumn{1}{c}{Task}\\ 
\cmidrule(lr){1-4}\cmidrule(lr){5-6}\cmidrule(lr){7-7}
\textit{AugMix} &\textit{Rand} &\textit{Trivial} &\textit{MaskOut} &\textit{LS} &\textit{DropOut} &MC\\
\midrule
\midrule
0.1 &0.1 &0.1 &0.1\textbar 3  &0.1 &0.5 &41.90\\
0.5 &0.5 &0.5 &0.7\textbar 3  &0.0 &0.5 &\textbf{44.91}\\
0.5 &0.5 &0.5 &0.7\textbar 10 &0.0 &0.5 &42.13\\
0.5 &0.5 &0.5 &0.7\textbar 3  &0.0 &0.6 &42.36\\
0.9 &0.9 &0.9 &0.7\textbar 3  &0.3 &0.7 &43.52\\
\bottomrule 
\end{tabular}
\begin{tablenotes}
\scriptsize
\item \textit{Rand}: \textit{RandAugment} \space  \textit{Trivial}: TrivialAugment \space \textit{LS}: \textit{Label Smoothing} \space MS: multiclass pain assessment. For Augmentation \& Regularization the first number represents the probability of application, while in \textit{MaskOut} the number followed \textbar \space indicates the number of square  sections applied.
\end{tablenotes}
\end{threeparttable}
\end{center}
\end{table}

\renewcommand{\arraystretch}{1.2}
\begin{table}
\caption{Classification results utilizing the facial video modality \& \textit{Concatenation} method, reported on accuracy \%.}
\label{table:videos_concat}
\begin{center}
\begin{threeparttable}
\begin{tabular}{ P{0.8cm}  P{0.8cm} P{0.8cm} P{1.1cm}  P{0.5cm}  P{1.1cm}  P{0.8cm} }
\toprule
\multicolumn{4}{c}{Augmentation} 
&\multicolumn{2}{c}{Regularization} 
&\multicolumn{1}{c}{Task}\\ 
\cmidrule(lr){1-4}\cmidrule(lr){5-6}\cmidrule(lr){7-7}
\textit{AugMix} &\textit{Rand} &\textit{Trivial} &\textit{MaskOut} &\textit{LS} &\textit{DropOut} &MC\\
\midrule
\midrule
0.3 &0.3 &0.3 &0.3\textbar 3 &0.1 &0.5 &40.28\\
0.5 &0.5 &0.5 &0.8\textbar 5 &0.0 &0.5 &41.44\\
0.9 &0.9 &0.9 &0.7\textbar 3 &0.2 &0.7 &42.13\\
0.9 &0.9 &0.9 &0.7\textbar 1 &0.4 &0.5 &41.90\\
0.9 &0.9 &0.9 &0.6\textbar 3 &0.4 &0.5 &\textbf{43.75}\\
\bottomrule 
\end{tabular}
\end{threeparttable}
\end{center}
\end{table}

\subsection{fNIRS}
Similar to facial videos, the methods of \textit{Addition} and \textit{Concatenation} were applied. From the original $24$ channels $2$ were excluded due to faults. For the HbR \& \textit{Addition} method, starting accuracy was $39.35\%$ with uniform probabilities of $0.5$ for \textit{AugMix}, \textit{Rand}, and \textit{Trivial}, and \textit{MaskOut} set to $0.6\vert5$. Adjustments to \textit{MaskOut} at $0.7\vert3$ and increased \textit{LS} led to slight accuracy dips, while further adjustments in \textit{LS} and \textit{DropOut} raised accuracy to $41.20\%$ (refer to Table \ref{table:hhb_addition}).
In the HbR \& \textit{Concatenation} method, initial augmentations with \textit{MaskOut} at $0.7\vert3$ achieved an accuracy of $40.97\%$. Amplifying all augmentations to $0.9$ while maintaining \textit{MaskOut} at $0.7\vert3$ resulted in a peak accuracy of $42.13\%$ (refer to Table \ref{table:hhb_concat}).
For the HbO \& \textit{Addition} method, accuracies began at $43.06\%$ with uniform augmentation probabilities of $0.3$ and \textit{MaskOut} at $0.3\vert3$. Raising \textit{MaskOut} to $0.7\vert3$ with slight changes in \textit{LS} and \textit{DropOut} maintained similar accuracies, while optimizing \textit{MaskOut} to $0.8\vert3$ improved performance to $44.68\%$ (refer to Table \ref{table:02hb_addition}).
In the HbO \& \textit{Concatenation} method, the augmentation methods with a probability of $0.1$ started with an accuracy of $42.13\%$. 
The peak accuracy of $44.44\%$ was achieved with a balanced augmentation at $0.9$ and \textit{MaskOut} at $0.7\vert3$, indicating effectiveness of increased overall applied augmentation combined with high regularization. Subsequent adjustments slightly lowered accuracy, underscoring the importance of optimal augmentation settings (refer to Table \ref{table:o2hb_concat}).Generally, enhanced performance is observed with HbO compared to HbR, as also noted in other studies \cite{rojas_huang_2017_c}, due to its superior signal-to-noise ratio.
The combined HbR and HbO using the \textit{Addition} method initially showed an accuracy of $42.82\%$ with all augmentations at zero except for \textit{MaskOut} at $0.7\vert3$. Increasing \textit{AugMix}, \textit{Rand}, and \textit{Trivial} to $0.5$ while elevating \textit{MaskOut} to $0.7\vert7$ marginally improved accuracy to $43.29\%$. Maintaining augmentations but adjusting \textit{MaskOut} back to $0.7\vert3$ with a slight increase in \textit{LS} resulted in a slight decrease in accuracy to $42.59\%$. However, further increasing all augmentations to $0.9$ and \textit{LS} to $0.3$ while maintaining \textit{MaskOut} at $0.7\vert3$ maximized the accuracy to $43.75\%$. Reducing \textit{DropOut} to $0.1$ in the final configuration slightly reduced accuracy to $43.06\%$, emphasizing the importance of optimizing regularization alongside augmentation strategies for achieving the best possible results (refer to Table \ref{table:hhb_o2hb_addition}).

\renewcommand{\arraystretch}{1.2}
\begin{table}
\caption{Classification results utilizing the HbR \& \textit{Addition} method, reported on accuracy \%.}
\label{table:hhb_addition}
\begin{center}
\begin{threeparttable}
\begin{tabular}{ P{0.8cm}  P{0.8cm} P{0.8cm} P{1.1cm}  P{0.5cm}  P{1.1cm}  P{0.8cm} }
\toprule
\multicolumn{4}{c}{Augmentation} 
&\multicolumn{2}{c}{Regularization} 
&\multicolumn{1}{c}{Task}\\ 
\cmidrule(lr){1-4}\cmidrule(lr){5-6}\cmidrule(lr){7-7}
\textit{AugMix} &\textit{Rand} &\textit{Trivial} &\textit{MaskOut} &\textit{LS} &\textit{DropOut} &MC\\
\midrule
\midrule
0.5 &0.5 &0.5 &0.6\textbar 5 &0.0 &0.5 &39.35\\
0.5 &0.5 &0.5 &0.7\textbar 3 &0.4 &0.5 &38.89\\
0.9 &0.9 &0.9 &0.7\textbar 3 &0.1 &0.9 &40.05\\
0.9 &0.9 &0.9 &0.7\textbar 5 &0.4 &0.5 &\textbf{41.20}\\
0.5 &0.5 &0.5 &0.7\textbar 3 &0.0 &0.4 &40.51\\
\bottomrule 
\end{tabular}
\end{threeparttable}
\end{center}
\end{table}

\renewcommand{\arraystretch}{1.2}
\begin{table}
\caption{Classification results utilizing the HbR \& \textit{Concatenation} method, reported on accuracy \%.}
\label{table:hhb_concat}
\begin{center}
\begin{threeparttable}
\begin{tabular}{ P{0.8cm}  P{0.8cm} P{0.8cm} P{1.1cm}  P{0.5cm}  P{1.1cm}  P{0.8cm} }
\toprule
\multicolumn{4}{c}{Augmentation} 
&\multicolumn{2}{c}{Regularization} 
&\multicolumn{1}{c}{Task}\\ 
\cmidrule(lr){1-4}\cmidrule(lr){5-6}\cmidrule(lr){7-7}
\textit{AugMix} &\textit{Rand} &\textit{Trivial} &\textit{MaskOut} &\textit{LS} &\textit{DropOut} &MC\\
\midrule
\midrule
0.0 &0.0 &0.7 &0.7\textbar 3 &0.0 &0.5 &40.97\\
0.5 &0.5 &0.5 &0.7\textbar 1 &0.0 &0.5 &41.44\\
0.9 &0.9 &0.9 &0.7\textbar 3 &0.1 &0.8 &\textbf{42.13}\\
0.9 &0.9 &0.9 &0.7\textbar 3 &0.4 &0.5 &41.20\\
0.5 &0.5 &0.5 &0.7\textbar 3 &0.0 &0.3 &39.81\\
\bottomrule 
\end{tabular}
\end{threeparttable}
\end{center}
\end{table}

\renewcommand{\arraystretch}{1.2}
\begin{table}
\caption{Classification results utilizing the HbO \& \textit{Addition} method, reported on accuracy \%.}
\label{table:02hb_addition}
\begin{center}
\begin{threeparttable}
\begin{tabular}{ P{0.8cm}  P{0.8cm} P{0.8cm} P{1.1cm}  P{0.5cm}  P{1.1cm}  P{0.8cm} }
\toprule
\multicolumn{4}{c}{Augmentation} 
&\multicolumn{2}{c}{Regularization} 
&\multicolumn{1}{c}{Task}\\ 
\cmidrule(lr){1-4}\cmidrule(lr){5-6}\cmidrule(lr){7-7}
\textit{AugMix} &\textit{Rand} &\textit{Trivial} &\textit{MaskOut} &\textit{LS} &\textit{DropOut} &MC\\
\midrule
\midrule
0.3 &0.3 &0.3 &0.3\textbar 3 &0.1 &0.5 &43.06\\
0.5 &0.5 &0.5 &0.7\textbar 3 &0.2 &0.5 &42.82\\
0.9 &0.9 &0.9 &0.7\textbar 3 &0.4 &0.8 &43.29\\
0.9 &0.9 &0.9 &0.7\textbar 9 &0.4 &0.5 &44.44\\
0.9 &0.9 &0.9 &0.8\textbar 3 &0.4 &0.5 &\textbf{44.68}\\
\bottomrule 
\end{tabular}
\end{threeparttable}
\end{center}
\end{table}

\renewcommand{\arraystretch}{1.2}
\begin{table}
\caption{Classification results utilizing the HbO \& \textit{Concatenation} method, reported on accuracy \%.}
\label{table:o2hb_concat}
\begin{center}
\begin{threeparttable}
\begin{tabular}{ P{0.8cm}  P{0.8cm} P{0.8cm} P{1.1cm}  P{0.5cm}  P{1.1cm}  P{0.8cm} }
\toprule
\multicolumn{4}{c}{Augmentation} 
&\multicolumn{2}{c}{Regularization} 
&\multicolumn{1}{c}{Task}\\ 
\cmidrule(lr){1-4}\cmidrule(lr){5-6}\cmidrule(lr){7-7}
\textit{AugMix} &\textit{Rand} &\textit{Trivial} &\textit{MaskOut} &\textit{LS} &\textit{DropOut} &MC\\
\midrule
\midrule
0.1 &0.1 &0.1 &0.1\textbar 3 &0.1 &0.5 &42.13\\
0.5 &0.5 &0.5 &0.0\textbar 0 &0.0 &0.5 &43.98\\ 
0.5 &0.5 &0.5 &0.7\textbar 1 &0.0 &0.5 &42.36\\ 
0.9 &0.9 &0.9 &0.7\textbar 3 &0.4 &0.9 &\textbf{44.44}\\ 
0.5 &0.5 &0.5 &0.7\textbar 3 &0.0 &0.8 &43.52\\
\bottomrule 
\end{tabular}
\end{threeparttable}
\end{center}
\end{table}

\renewcommand{\arraystretch}{1.2}
\begin{table}
\caption{Classification results utilizing the HbR, HbO \& \textit{Addition} method, reported on accuracy \%.}
\label{table:hhb_o2hb_addition}
\begin{center}
\begin{threeparttable}
\begin{tabular}{ P{0.8cm}  P{0.8cm} P{0.8cm} P{1.1cm}  P{0.5cm}  P{1.1cm}  P{0.8cm} }
\toprule
\multicolumn{4}{c}{Augmentation} 
&\multicolumn{2}{c}{Regularization} 
&\multicolumn{1}{c}{Task}\\ 
\cmidrule(lr){1-4}\cmidrule(lr){5-6}\cmidrule(lr){7-7}
\textit{AugMix} &\textit{Rand} &\textit{Trivial} &\textit{MaskOut} &\textit{LS} &\textit{DropOut} &MC\\
\midrule
\midrule
0.0 &0.0 &0.7 &0.7\textbar 3 &0.0 &0.5 &42.82\\
0.5 &0.5 &0.5 &0.7\textbar 7 &0.0 &0.5 &43.29\\
0.5 &0.5 &0.5 &0.7\textbar 3 &0.1 &0.5 &42.59\\
0.9 &0.9 &0.9 &0.7\textbar 3 &0.3 &0.9 &\textbf{43.75}\\
0.5 &0.5 &0.5 &0.7\textbar 3 &0.0 &0.1 &43.06\\
\bottomrule 
\end{tabular}
\end{threeparttable}
\end{center}
\end{table}

\renewcommand{\arraystretch}{1.2}
\begin{table}
\caption{Classification results utilizing the facial videos, HbO \& \textit{Addition} method, reported on accuracy \%.}
\label{table:video_o2hb_addition}
\begin{center}
\begin{threeparttable}
\begin{tabular}{ P{0.8cm}  P{0.8cm} P{0.8cm} P{1.1cm}  P{0.5cm}  P{1.1cm}  P{0.8cm} }
\toprule
\multicolumn{4}{c}{Augmentation} 
&\multicolumn{2}{c}{Regularization} 
&\multicolumn{1}{c}{Task}\\ 
\cmidrule(lr){1-4}\cmidrule(lr){5-6}\cmidrule(lr){7-7}
\textit{AugMix} &\textit{Rand} &\textit{Trivial} &\textit{MaskOut} &\textit{LS} &\textit{DropOut} &MC\\
\midrule
\midrule
0.5 &0.5 &0.5 &0.4\textbar 5 &0.0 &0.5 &42.36\\
0.5 &0.5 &0.5 &0.7\textbar 9 &0.0 &0.5 &41.67\\
0.9 &0.9 &09. &0.7\textbar 3 &0.1 &0.6 &42.59\\
0.9 &0.9 &0.9 &0.7\textbar 3 &0.3 &0.9 &43.06\\\
0.9 &0.9 &0.9 &0.7\textbar 5 &0.4 &0.5 &\textbf{43.75}\\
\bottomrule 
\end{tabular}
\end{threeparttable}
\end{center}
\end{table}

\renewcommand{\arraystretch}{1.2}
\begin{table}
\caption{Classification results utilizing the facial videos, HbO \& \textit{Single Diagram} method, reported on accuracy \%.}
\label{table:video_o2hb_1_image}
\begin{center}
\begin{threeparttable}
\begin{tabular}{ P{0.8cm}  P{0.8cm} P{0.8cm} P{1.1cm}  P{0.5cm}  P{1.1cm}  P{0.8cm} }
\toprule
\multicolumn{4}{c}{Augmentation} 
&\multicolumn{2}{c}{Regularization} 
&\multicolumn{1}{c}{Task}\\ 
\cmidrule(lr){1-4}\cmidrule(lr){5-6}\cmidrule(lr){7-7}
\textit{AugMix} &\textit{Rand} &\textit{Trivial} &\textit{MaskOut} &\textit{LS} &\textit{DropOut} &MC\\
\midrule
\midrule
0.5 &0.5 &0.5 &0.3\textbar 5 &0.0 &0.5 &45.83\\
0.9 &0.9 &0.9 &0.7\textbar 3 &0.1 &0.6 &\textbf{46.76}\\
0.9 &0.9 &0.9 &0.7\textbar 3 &0.3 &0.6 &46.53\\
0.9 &0.9 &0.9 &0.9\textbar 3 &0.4 &0.5 &45.83\\\
0.5 &0.5 &0.5 &0.7\textbar 3 &0.0 &0.7 &45.14\\
\bottomrule 
\end{tabular}
\end{threeparttable}
\end{center}
\end{table}

\subsection{Fusion}
In this section, we describe the fusion of facial videos and fNIRS. The HbO was utilized solely for the experiments since it demonstrated superior performance to the HbR. 
Two methods were developed for data fusion: the previously described \textit{Addition} method, aggregating embeddings from video frames and fNIRS channels and then combines them, and the \textit{Single-Diagram} method, where aggregated embeddings from both modalities are concurrently visualized in the same image.
For the \textit{Addition} method and initial configurations with moderate augmentation levels ($0.5$ for \textit{AugMix}, \textit{Rand}, \textit{Trivial}) and \textit{MaskOut} at $0.4\vert5$ achieved a $42.36\%$ accuracy. 
Increasing augmentation levels to $0.9$ and adjusting regularization parameters (\textit{LS} up to $0.4$ and \textit{DropOut} up to $0.9$) improved the accuracy, peaking at $43.75\%$ (refer to Table \ref{table:video_o2hb_addition}).
For the \textit{Single Diagram} method, accuracy improvements were observed, as shown in Table \ref{table:video_o2hb_1_image}. Starting with lower \textit{MaskOut} levels at $0.3\vert5$ and standard augmentation probabilities ($0.5$), the accuracy was $45.83\%$. Utilizing augmentation probabilities to $0.9$ and \textit{MaskOut} adjustments to $0.7\vert3$ significantly improved performance, achieving a high of $46.76\%$.

\section{Interpretation \& Comparison}
\begin{figure}
\begin{center}
\includegraphics[scale=0.275]{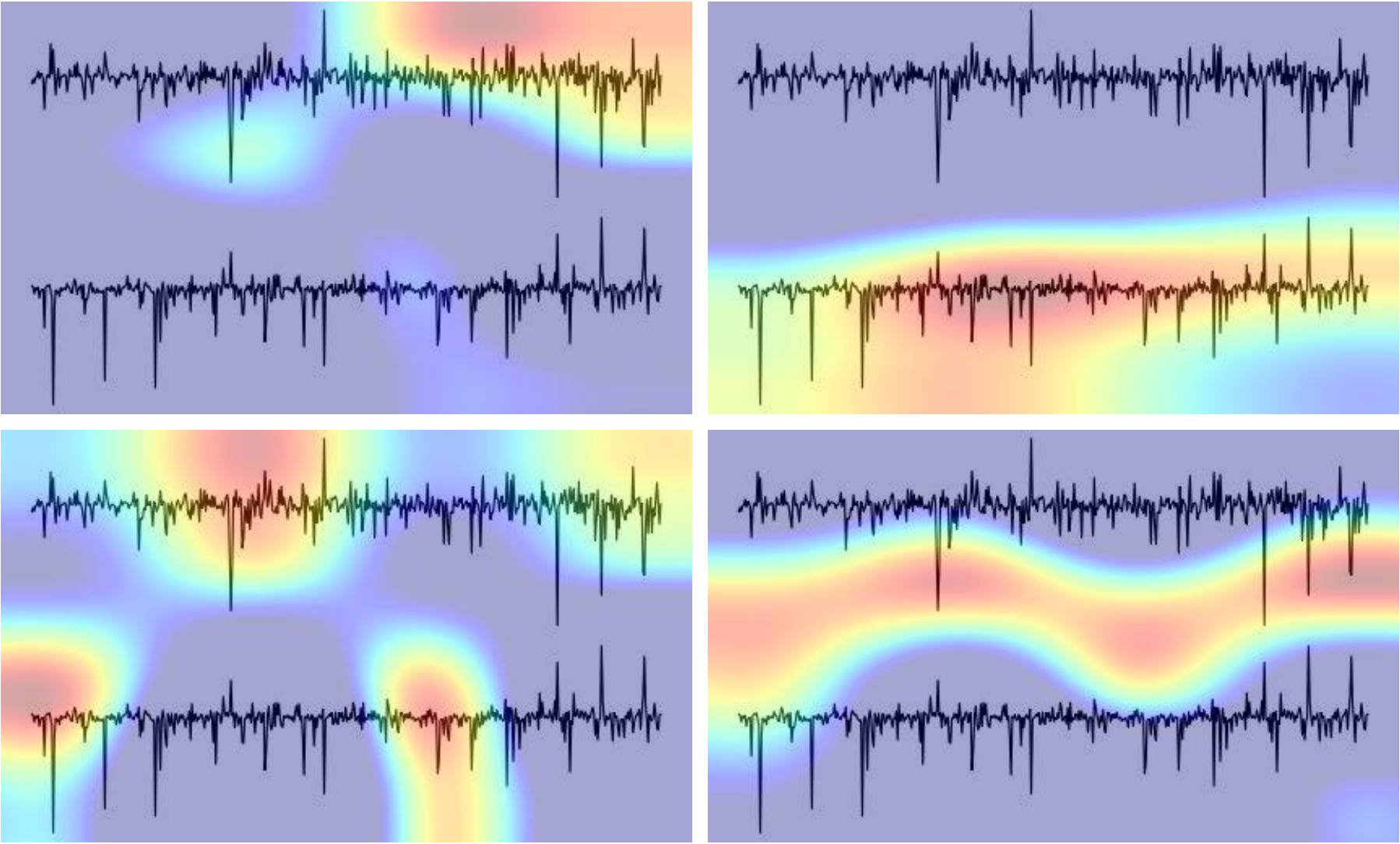}
\end{center}
\caption{Attention maps from the \textit{PainViT--2}.}
\label{attention}
\end{figure}

Regarding the framework's interpretation, attention maps were generated from the last layer of \textit{PainViT--2}, which processed the unified image visualizing both the video and HbO embedding waveforms. This layer contains $500$ neurons, each contributing uniquely to and attending to the input. Fig. \ref{attention} illustrates four examples where certain neurons focus on the video embedding waveform, others on the HbO, and some attend to both waveforms, emphasizing different parts and details. 
Table \ref{table:comparison} compares the proposed pipeline and the baseline results provided by the challenge organizers. The video-based approach using the \textit{Addition} method outperformed the baseline by $4.91\%$. Using the HbO with the \textit{Addition} method, the improvement was lesser,  at $1.48\%$. Finally, the modality fusion using the \textit{Single Diagram} approach resulted in a more significant improvement of $6.56\%$.

\renewcommand{\arraystretch}{1.2}
\begin{table}
\caption{Comparison with the validation baseline provided by the \textit{AI4PAIN} challenge organizers, reported on accuracy \%.}
\label{table:comparison}
\begin{center}
\begin{threeparttable}
\begin{tabular}{ P{1.8cm} P{1.8cm} P{1.8cm} P{1.8cm}}
\toprule
\multirow{2}[2]{*}{\shortstack{Approach}}
&\multicolumn{3}{c}{Modality}\\ 
\cmidrule(lr){2-4}
&Video &fNIRS &Fusion\\
\midrule
\midrule
Baseline &40.00 &43.20 &40.20\\
Our &44.91 &44.68 &46.76\\
        
\bottomrule 
\end{tabular}
\begin{tablenotes}
\scriptsize
\item 
\end{tablenotes}
\end{threeparttable}
\end{center}
\end{table}

\section{Conclusion}
This study outlines our contribution to the \textit{First Multimodal Sensing Grand Challenge for Next-Gen Pain Assessment (AI4PAIN)}, employing facial videos and fNIRS through a modality-agnostic approach. 
\textit{Twins-PainViT} was introduced, a framework founded on a dual configuration of vision transformers, pre-trained on the plethora of datasets in a multi-task learning setting. 
Furthermore, a fundamental component of the proposed pipeline was the waveform representation, which was applied to the original fNIRS data and the learned embeddings from both modalities. This approach of extracting embeddings and integrating them into a single image diagram effectively and efficiently eliminated the need for dedicated domain-specific models for each modality.
The conducted experiments showcased high performances for unimodal and multimodal settings, surpassing the provided baseline results. 
Additionally, the interpretation of \textit{Pain-ViT--2} through the creation of attention maps for the image diagrams showed that specific neurons target particular modalities or distinct aspects of them, indicating a holistic consideration in the analysis process.
We suggest that future research employ multimodal approaches, which have proven to be the most effective method for assessing pain in real-world settings. It is also essential to develop methods for interpreting data, particularly to facilitate the integration of these frameworks into clinical practice.

\section*{Ethical Impact Statement}
This research employed the \textit{AI4PAIN} dataset  \cite{ai4pain,rojas_hirachan_2023} provided by the challenge organizers to evaluate the proposed methods.
The participants did not report any prior history of neurological or psychiatric disorders, current unstable medical conditions, chronic pain, or regular medication use at the time of testing. Upon arrival, they received a detailed explanation of the experimental procedures. Written informed consent was obtained before the experiment began. The experimental procedures involving human subjects described in the original paper were approved by the University of Canberra's Human Ethics Committee (\textit{approval number: 11837}).
This study presents a pain assessment framework for continuous patient monitoring and minimizing human biases. However, it is crucial to acknowledge that deploying this framework in real-world clinical environments may pose challenges, requiring additional experiments and thorough validation through clinical trials prior to final deployment.
Furthermore, the sole facial image featured in this study is a designed illustration and does not represent an actual person.

In addition, several datasets were utilized to pretrain the proposed pain assessment framework. 
The \textit{AffectNet} \cite{mollahosseini_hasani_2019} dataset is compiled using search engine queries. The original paper does not explicitly detail ethical compliance measures. 
The \textit{RAF-DB} \cite{li_deng_2017} dataset was compiled using the Flickr image hosting service. Although Flickr hosts both public and privately shared images, the authors do not explicitly mention the type of the downloaded images. 
The original paper of \textit{Compound FEE-DB} \cite{du_tao_2014} does not mention ethical compliance measures, but only that the subjects were recruited from the Ohio State University area and received a monetary reward for participating.
The \textit{EEG-BST-SZ} \cite{ford_2013} dataset was recorded with assistance from trained research assistants, psychiatrists, or clinical psychologists who conducted all interviews. The study received approval from the University of California at San Francisco Institutional Review Board and the San Francisco Veterans Affairs Medical Center. 
The original paper on the \textit{Silent-EMG} \cite{gaddy_klein_2020} dataset does not explicitly mention adherence to ethical compliance measures. 
However, it is noted that the data recorded came solely from one individual, also one of the authors and creators of the dataset.
The data from the \textit{BioVid Heat Pain Database} \cite{biovid_2013} were recorded according to the ethical guidelines of Helsinki (\textit{ethics committee: 196/10-UBB/bal}).

\section*{Acknowledgement}
Research supported by the \textit{ODIN} project that has received funding from
the European Union’s Horizon 2020 research and innovation program under
grant agreement No $101017331$.

\bibliographystyle{IEEEtran}
\bibliography{library}

\appendix
\subsection*{Supplementary Metrics}

\renewcommand{\arraystretch}{1.2}
\begin{table}[H]
\caption{Classification results of the proposed approaches, reported on macro-averaged precision, recall and F1 score.}
\label{table:comparison}
\begin{center}
\begin{threeparttable}
\begin{tabular}{ P{1.0cm} P{2.5cm} P{1.0cm} P{1.0cm} P{1.0cm}}
\toprule
\multirow{2}[2]{*}{\shortstack{Modality}}
&\multirow{2}[2]{*}{\shortstack{Approach}}
&\multicolumn{3}{c}{Metrics}\\ 
\cmidrule(lr){3-5}
& &Precision &Recall &F1\\
\midrule
\midrule
Video  &\textit{Addition} &44.91 &44.97 &44.60\\
fNIRS  &HbO \& \textit{Addition} &44.68 &45.08 &43.60\\
Fusion &\textit{Single Diagram} &46.76 &47.29 &46.70\\
        
\bottomrule 
\end{tabular}
\begin{tablenotes}
\scriptsize
\item 
\end{tablenotes}
\end{threeparttable}
\end{center}
\end{table}

\subsection*{Supplementary Attention Maps}
\label{additional_attention_maps}

\begin{figure}[h!]
\begin{center}
\includegraphics[scale=0.275]{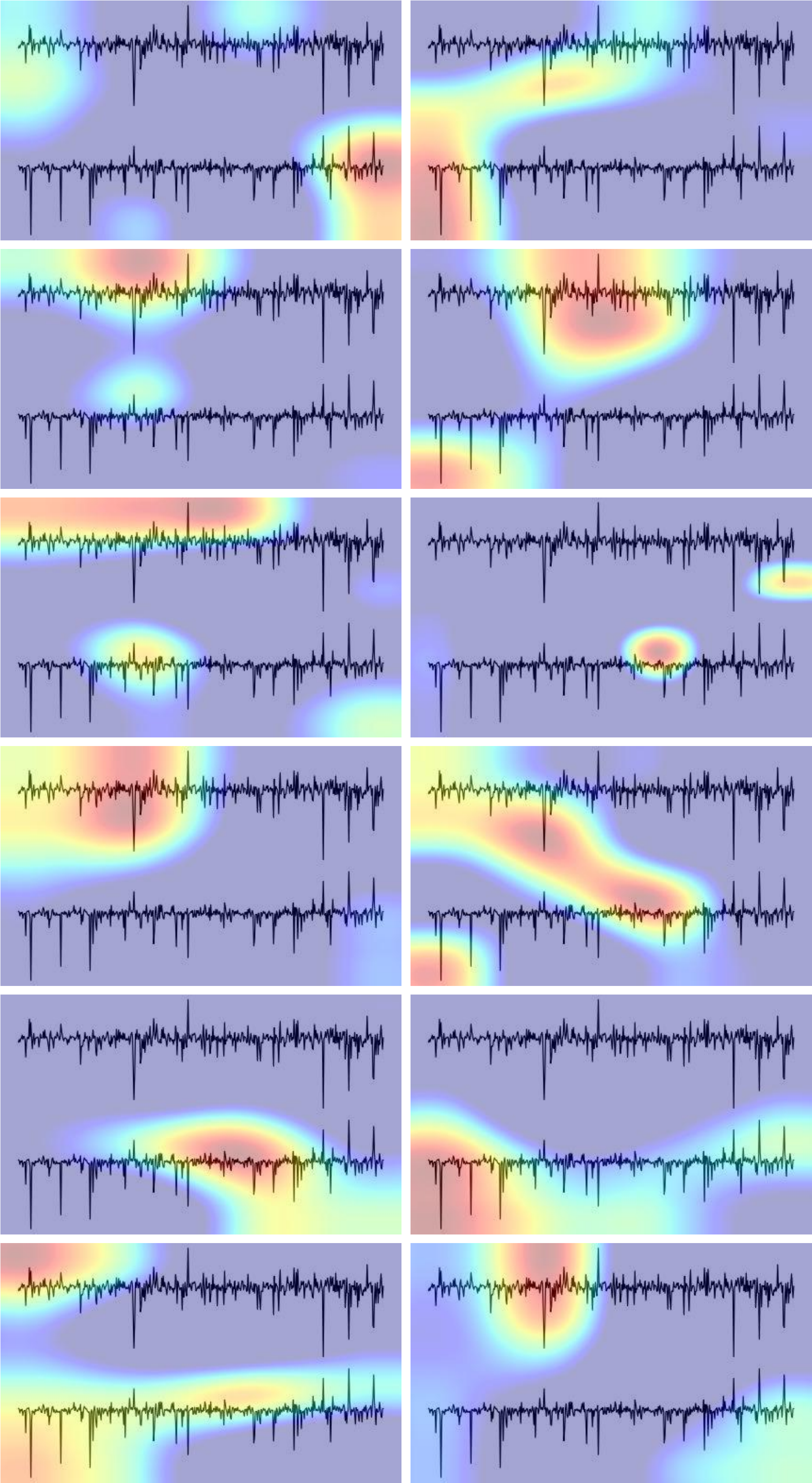} 
\end{center}
\caption{Additional attention maps from the \textit{PainViT--2}.}
\label{attention_1}
\end{figure}

%
%
%
%
%
%
%
%
%
%

\end{document}